# Multiagent Bidirectionally-Coordinated Nets

## Emergence of Human-level Coordination in Learning to Play StarCraft Combat Games[*]

**Peng Peng[♮], Ying Wen[§†], Yaodong Yang[§], Yuan Quan[♮], Zhenkun Tang[♮], Haitao Long[♮], Jun Wang[§]**

[§]University College London, [♮]Alibaba Group

## Abstract

Many artificial intelligence (AI) applications often require multiple intelligent agents to work in a collaborative effort. Efficient learning for intra-agent communication and coordination is an indispensable step towards general AI. In this paper, we take StarCraft combat game as a case study, where the task is to coordinate multiple agents as a team to defeat their enemies. To maintain a scalable yet effective communication protocol, we introduce a Multiagent Bidirectionally-Coordinated Network (BiCNet ['bɪknet]) with a vectorised extension of actor-critic formulation. We show that BiCNet can handle different types of combats with arbitrary numbers of AI agents for both sides. Our analysis demonstrates that without any supervisions such as human demonstrations or labelled data, BiCNet could learn various types of advanced coordination strategies that have been commonly used by experienced game players. In our experiments, we evaluate our approach against multiple baselines under different scenarios; it shows state-of-the-art performance, and possesses potential values for large-scale real-world applications.

## Introduction

The last decade has witnessed massive progresses in the field of Artificial Intelligence (AI). With supervision from labelled data, machines have, to some extent, exceeded human-level perception on visual recognitions and speech recognitions, while fed with feedback reward, single AI units (*aka* agents) defeat humans in various games including Atari video games (Mnih et al. 2015), Go game (Silver et al. 2016), and card game (Brown and Sandholm 2017).

Yet, true human intelligence embraces social and collective wisdom which lays an essential foundation for reaching the grand goal of Artificial General Intelligence (AGI) (Goertzel and Pennachin 2007). As demonstrated by crowd sourcing, aggregating efforts collectively from the public would solve the problem that otherwise is unthinkable by a single person. Even social animals like a brood of well-organised ants could accomplish challenging tasks such as hunting, building a kingdom, and even waging a war, although each ant by itself is weak and limited. Interestingly, in the coming era of algorithmic economy, AI agents with a certain rudimentary level of *artificial* collective intelligence start to emerge from multiple domains. Typical examples include the trading robots gaming on the stock markets (Deboeck 1994), ad bidding agents competing with each other over online advertising exchanges (Wang, Zhang, and Yuan 2017), and e-commerce collaborative filtering recommenders predicting user interests through the wisdom of the crowd (Schafer, Konstan, and Riedl 1999).

We thus believe a next grand challenge of AGI is to answer how multiple AI agents could learn human-level collaborations, or competitions, from their experiences with the environment where both of their incentives and economic constraints co-exist. As the flourishes of deep reinforcement learning (DRL) (Mnih et al. 2015; Silver et al. 2016), researchers start to shed light on tackling multiagent problems (Schmidhuber 1996) with the enhanced learning capabilities, e.g., (Sukhbaatar, Fergus, and others 2016; Mordatch and Abbeel 2017).

In this paper, we leverage a real-time strategy game, *StarCraft*[1], as the use case to explore the learning of intelligent collaborative behaviours among multiple agents. Particularly, we focus on StarCraft micromanagement tasks (Synnaeve et al. 2016), where each player controls their own units (with different functions to collaborate) to destroy the opponents army in the combats under different terrain conditions. Such game is considered as one of the most difficult games for computers with more possible states than Go game (Synnaeve et al. 2016). The learning of this large-scale multiagent system faces a major challenge that the parameters space grows exponentially with the increasing number of agents involved. As such, the behaviours of the agents can become so sophisticated that any joint learner method (Sukhbaatar, Fergus, and others 2016) would be inefficient and unable to deal with the changing number of agents in the game.

We formulate multiagent learning for StarCraft combat tasks as a zero-sum Stochastic Game. Agents are communicated by our proposed bidirectionally-coordinated net (BiCNet), while the learning is done using a multiagent actor-critic framework. In addition, we also introduce parameter sharing to solve the scalability issue. We observe that BiCNet can automatically learn various optimal strategies to coordinating agents, similar to what experienced human players would adopt in playing the StarCraft game, ranging from trivial *move without collision* to a basic tactic *hit and run* to sophisticated *cover attack*, and *focus fire without overkill*. We have conducted our experiments by testing over a set of combat tasks with different levels of difficulties. Our method outperforms state-of-the-art methods and shows its potential usage in a wide range of multiagent tasks in the real-world applications.

---

[*]Previously as title: "Multiagent Bidirectionally-Coordinated Nets for Learning to Play StarCraft Combat Games", Mar 2017.

[†]The first two authors have equal contributions. Correspondence to Jun Wang, jun.wang@cs.ucl.ac.uk.

[1]Trademark of Blizzard Entertainment[TM].

## Related Work

The studies on interaction and collaboration in multiagent settings have a long history (Littman 1994; Schmidhuber 1996). Although limited to toy examples in the beginning, reinforcement learning, as a means, has long been applied to multiagent systems in order to learn optimal collaboration policies. One of the key components in multiagent RL is to learn a communication protocol among agents. With deep learning, representative solutions include the differentiable inter-agent learning (DIAL) (Foerster et al. 2016) and the CommNet (Sukhbaatar, Fergus, and others 2016), both of which are end-to-end trainable by back-propagation.

DIAL (Foerster et al. 2016) was introduced in partially observable settings where messages passing between agents are allowed. The agent is also named as a *independent learner*. The idea of learning independent agents can also be found (Lauer and Riedmiller 2000; Kapetanakis and Kudenko 2002; Lauer and Riedmiller 2004; Foerster et al. 2016). In DIAL, each agent consists of a recurrent neural network that outputs individual agent's Q-value and a message to transfer for each time-step. The generated messages is then transferred to other agents as used as inputs for others in the next time step. The received messages will be embedded with agent's current observations and last action as the representation of the global information. Communication between independent agents is one way to mitigate the notorious non-stationary issue in the mutliagent settings as the gradients will at least flow among the agents; however, researchers are still looking for better solutions for complex environments such as StarCraft.

By contrast, CommNet (Sukhbaatar, Fergus, and others 2016) is designed for *joint action learners* in fully observable settings. Unlike DIAL, CommNet proposes a single network in the multiagent setting, passing the averaged message over the agent modules between layers. However, as the communication network is fully symmetric and embedded in the original network, it lacks the ability of handle heterogeneous agent types. Also it is a single network for all agents, and therefore its scalability is unclear. In this paper, we solve these issues by creating a dedicated bi-directional communication channel using recurrent neural networks (Schuster and Paliwal 1997). As such, heterogeneous agents can be created with a different set of parameters and output actions. The bi-directional nature means that the communication is not entirely symmetric, and the different priority among agents would help solving any possible tie between multiple optimal joint actions (Busoniu, Babuska, and De Schutter 2008; Spaan et al. 2002).

Multiagent systems have been explored on specific StarCraft games. Google DeepMind released a game interface based on StarCraft II and claimed that it is hard to make significant progress on the full game even with the state-of-the-art RL algorithms (Vinyals et al. 2017). Usunier et al. presented a heuristic exploration technique for learning deterministic policies in micro-management tasks. Both Synnaeve et al. and Usunier et al. focused on a greedy MDP approach, i.e., the action of an agent is dependent explicitly on the action of other agents. In our paper, the dependency of agents is rather modelled over hidden layers by making use of bi-directional RNN (Schuster and Paliwal 1997). A significant benefit over the greedy solution is that, while keeping simple, the communication happens in the latent space so that high-level information can be passed between agents; meanwhile, the gradient updates from all the agents can be efficiently propagated through the entire network.

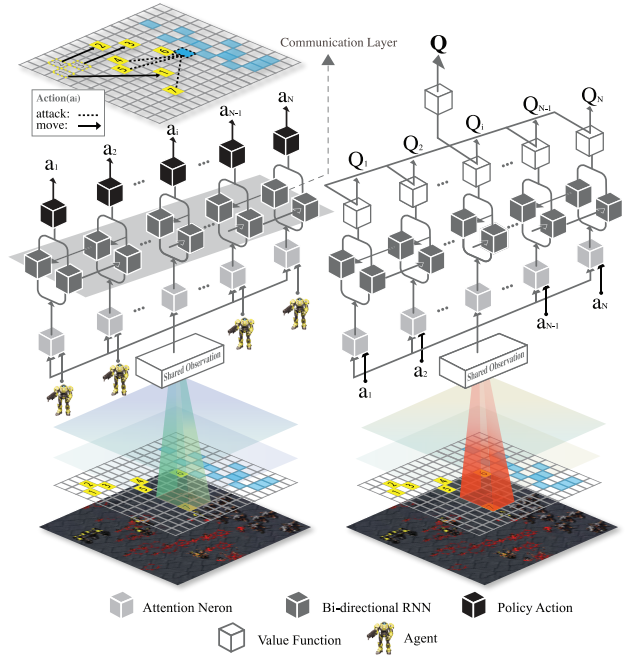

(a) Multiagent policy networks  (b) Multiagent Q networks

Figure 1: Bidirectionally-Coordinated Net (BiCNet). As a means of communication, bi-direction recurrent networks have been used to connect each individual agent's policy and and Q networks. The learning is done by multiagent deterministic actor-critic as derived in the text.

Recently, Foerster et al. has attempted to solve the non-stationarity problem in mutliagent learning by improving the replay buffer, and tested the DIAL model in a way that all agents are fully decentralised. The COMA model (Foerster et al. 2017a) was then proposed to tackle the challenge of multiagent credit assignment. Through the introduction of the counterfactual reward; the idea of training multiagent systems in the centralised critic and decentralised actors way was further reinforced. At the same time, the framework of centralised learning and decentralised execution was also adopted by MADDPG in (Lowe et al. 2017) in some simpler, non-startcraft cases. By contrast, our BiCNet makes use of memory to form a communication channel among agents where the parameter space of communication is independent of the number of agents.

## Multiagent Bidirectionally-Coordinated Nets

### StartCraft Combat as Stochastic Games

The StarCraft combat games, *a.k.a.*, the micromanagement tasks, refer to the low-level, short-term control of the army members during a combat against the enemy members. For each combat, the agents in one side are fully cooperative, and they compete with the opponents; therefore, each combat can be considered as a zero-sum competitive game between $N$ agents and $M$ enemies. We formulate it as a zero-sum Stochastic Game (SG) (Owen 1995), *i.e.*, a dynamic game in a multiple state situation played by multiple agents. A SG can be described by a tuple $(\mathcal{S}, \{\mathcal{A}_i\}_{i=1}^{N}, \{\mathcal{B}_i\}_{i=1}^{M}, \mathcal{T}, \{\mathcal{R}_i\}_{i=1}^{N+M})$. Let $S$ denotes the state space of the current game, shared among all the agents. Initial state $\mathbf{s}_1$ follows $\mathbf{s}_1 \sim p_1(\mathbf{s})$. We define the action space of the controlled agent $i$ as $\mathcal{A}_i$, and the action space of the enemy $j$ as $\mathcal{B}_j$. $\mathcal{T} : \mathcal{S} \times \mathcal{A}^N \times \mathcal{B}^M \to \mathcal{S}$ stands for the deterministic transition function of the environment, and $\mathcal{R}_i : \mathcal{S} \times \mathcal{A}^N \times \mathcal{B}^M \to \mathbb{R}$ represents the reward function of each agent $i$ for $i \in [1, N]$.

In order to maintain a flexible framework that could allow an arbitrary number of agents, we consider that the agents, either the controlled or the enemies, share the same state space $S$ of the current game; and within each camp, agents are homogeneous[2] thus having the same action spaces $\mathcal{A}$ and $\mathcal{B}$ respectively. That is, for each agent $i \in [1, N]$ and enemy $j \in [1, M]$, $\mathcal{A}_i = \mathcal{A}$ and $\mathcal{B}_j = \mathcal{B}$. As discrete action space is intractably large, we consider continuous control outputs, e.g., attack angle and distance.

In defining the reward function, we introduce a time-variant global reward based on the difference of the heath level between two consecutive time steps:

$$r(\mathbf{s}, \mathbf{a}, \mathbf{b}) \equiv \frac{1}{M} \sum_{j=N+1}^{N+M} \Delta \mathcal{R}_j^t(\mathbf{s}, \mathbf{a}, \mathbf{b}) - \frac{1}{N} \sum_{i=1}^{N} \Delta \mathcal{R}_i^t(\mathbf{s}, \mathbf{a}, \mathbf{b}), \quad (1)$$

where for simplicity, we drop the subscript $t$ in global reward $r(\mathbf{s}, \mathbf{a}, \mathbf{b})$. For given time step $t$ with state $\mathbf{s}$, the controlled agents take actions $\mathbf{a} \in \mathcal{A}^N$, the opponents take actions $\mathbf{b} \in \mathcal{B}^M$, and $\Delta \mathcal{R}_i^t(\cdot) \equiv \mathcal{R}_i^{t-1}(\mathbf{s}, \mathbf{a}, \mathbf{b}) - \mathcal{R}_i^t(\mathbf{s}, \mathbf{a}, \mathbf{b})$ represents the reduced health level for agent $i$. Note that Eq.(1) is presented from the aspect of controlled agents; the enemy's global reward is the exact opposite, making the sum of rewards from both camps equal zero. As the health level is non-increasing over time, Eq. (1) gives a positive reward at time step $t$ if the decrease of enemies' health levels exceeds that of ours.

With the defined global reward $r(\mathbf{s}, \mathbf{a}, \mathbf{b})$, the controlled agents jointly take actions $\mathbf{a}$ in state $\mathbf{s}$ when the enemies take joint actions $\mathbf{b}$. The agents' objective is to learn a policy that *maximises* the expected sum of discounted rewards, i.e., $\mathbb{E}[\sum_{k=0}^{+\infty} \lambda^k r_{t+k}]$, where $0 \le \lambda < 1$ is discount factor. Conversely, the enemies' joint policy is to *minimise* the expected sum. Correspondingly, we have the following *Minimax* game:

$$Q_{\text{SG}}^*(\mathbf{s}, \mathbf{a}, \mathbf{b}) = r(\mathbf{s}, \mathbf{a}, \mathbf{b}) + \lambda \max_\theta \min_\phi Q_{\text{SG}}^*(\mathbf{s}', \mathbf{a}_\theta(\mathbf{s}'), \mathbf{b}_\phi(\mathbf{s}')), \quad (2)$$

where $\mathbf{s}' \equiv \mathbf{s}^{t+1}$ is determined by $\mathcal{T}(\mathbf{s}, \mathbf{a}, \mathbf{b})$. $Q_{\text{SG}}^*(\mathbf{s}, \mathbf{a}, \mathbf{b})$ is the optimal action-state value function, which follows the Bellman Optimal Equation. Here we propose to use deterministic policy $\mathbf{a}_\theta : \mathcal{S} \to \mathcal{A}^N$ of the controlled agents and the deterministic policy (Silver et al. 2014) $\mathbf{b}_\phi : \mathcal{S} \to \mathcal{B}^M$ of the enemies. In small-scale MARL problems, a common solution is to employ *Minimax Q-learning* (Littman 1994). However, minimax Q-learning is generally intractable to apply in complex games. For simplicity, we consider the case that the policy of enemies is fixed, while leaving dedicated opponent modelling for future work. Then, SG defined in Eq. (2) effectively turns into an MDP problem (He et al. 2016):

$$Q^*(\mathbf{s}, \mathbf{a}) = r(\mathbf{s}, \mathbf{a}) + \lambda \max_\theta Q^*(\mathbf{s}', \mathbf{a}_\theta(\mathbf{s}')), \quad (3)$$

where we drop notation $\mathbf{b}_\phi$ for brevity.

---

[2] With our framework heterogeneous agents can be also trained using different parameters and action space.

**Local, Individual Rewards**

A potential drawback of only using the global reward in Eq. (1) and its resulting zero-sum game is that it ignores the fact that a team collaboration typically consists of local collaborations and reward function and would normally includes certain internal structure. Moreover, in practice, each agent tends to have its own objective which drives the collaboration. To model this, we extend the formulation in the previous section by replacing Eq. (1) with each agent's local reward and including the evaluation of their attribution to other agents that have been interacting with (either completing or collaborating), i.e.,

$$r_i(\mathbf{s}, \mathbf{a}, \mathbf{b}) \equiv \frac{1}{|top\text{-}K\text{-}u(i)|} \sum_{j \in top\text{-}K\text{-}u(i)} \Delta \mathcal{R}_j^t(\mathbf{s}, \mathbf{a}, \mathbf{b}) - \frac{1}{|top\text{-}K\text{-}e(i)|} \sum_{i' \in top\text{-}K\text{-}e(i)} \Delta \mathcal{R}_{i'}^t(\mathbf{s}, \mathbf{a}, \mathbf{b}), \quad (4)$$

where each agent $i$ maintains *top-K-u(i)* and *top-K-e(i)*, the top-$K$ lists of other agents and enemies, that are currently being interacted with. Replacing it with Eq. (1), we have $N$ number of Bellman equations for agent $i$, where $i \in \{1, ..., N\}$, for the same parameter $\theta$ of the policy function:

$$Q_i^*(\mathbf{s}, \mathbf{a}) = r_i(\mathbf{s}, \mathbf{a}) + \lambda \max_\theta Q_i^*(\mathbf{s}', \mathbf{a}_\theta(\mathbf{s}')). \quad (5)$$

**Communication w/ Bidirectional Backpropagation**

Although Eq. (5) makes single-agent methods, such as deterministic policy gradient (Silver et al. 2014; Mnih et al. 2016), immediately applicable for learning individual actions, those approaches, however, lacks a principled mechanism to foster team-level collaboration. In this paper, we allow communications between agents (right before taking individual actions) by proposing a bidirectionally-coordinated net (BiCNet).

Overall, BiCNet consists of a multiagent actor network and a multiagent critic network as illustrated in Fig.(1). Both of the policy network (actor) and the Q-network (critic) are based on the bi-directional RNN structure (Schuster and Paliwal 1997). The policy network, which takes in a shared observation together with a local view, returns the action for each individual agent. As the bi-directional recurrent structure could serve not only as a communication channel but also as a local memory saver, each individual agent is able to maintain its own internal states, as well as to share the information with its collaborators.

For the learning over BiCNet, intuitively, we can think of computing the backward gradients by unfolding the network of length $N$ (the number of controlled agents) and then applying backpropagation through time (BPTT) (Werbos 1990).

The gradients pass to both the individual $Q_i$ function and the policy function. They are aggregated from all the agents and their actions. In other words, the gradients from all agents rewards are first propagated to influence each of agents actions, and the resulting gradients are further propagated back to updating the parameters.

To see this mathematically, we denote the objective of a single agent $i$ by $J_i(\theta)$; that is to maximise its expected cumulative individual reward $r_i$ as $J_i(\theta) = \mathbb{E}_{\mathbf{s} \sim \rho^{\mathcal{T}}_{\mathbf{a}_\theta}}[r_i(\mathbf{s}, \mathbf{a}_\theta(\mathbf{s}))]$, where $\rho^{\mathcal{T}}_{\mathbf{a}_\theta}(\mathbf{s})$ is the discounted state distribution corresponding to the policy $\mathbf{a}_\theta$ under the transition $\mathcal{T}$, i.e., $\rho^{\mathcal{T}}_{\mathbf{a}_\theta}(\mathbf{s}) := \int_{\mathcal{S}} \sum_{t=1}^{\infty} \lambda^{t-1} p_1(\mathbf{s}) \mathbb{1}(\mathbf{s}' = \mathcal{T}^1_{\mathbf{a}_\theta, \mathbf{b}_\phi}(\mathbf{s})) d\mathbf{s}$; it can also be chosen as the stationary distribution of an ergodic MDP. So, we can write the objective of $N$ agents denoted by $J(\theta)$ as follows:

$$J(\theta) = \mathbb{E}_{\mathbf{s} \sim \rho^{\mathcal{T}}_{\mathbf{a}_\theta}}\left[\sum_{i=1}^{N} r_i(s, \mathbf{a}_\theta(\mathbf{s}))\right]. \quad (6)$$

Next, we introduce a multiagent analogue to the deterministic policy gradient theorem. The proof, which we give in the *Supplementary Material*, follows a similar scheme to both (Silver et al. 2014) and (Sutton et al. 2000).

**Theorem 1 (Multiagent Deterministic PG Theorem)**
*Given $N$ agents which are collectively represented in a policy parameterised with $\theta$, the discounted state distribution $\rho^{\mathcal{T}}_{\mathbf{a}_\theta}(\mathbf{s})$, and the objective function $J(\theta)$ defined in Eq.(6), we have the policy gradient as follows:*

$$\nabla_\theta J(\theta) = \mathbb{E}_{\mathbf{s} \sim \rho^{\mathcal{T}}_{\mathbf{a}_\theta}(\mathbf{s})}\left[\sum_{i=1}^{N}\sum_{j=1}^{N} \nabla_\theta \mathbf{a}_{j,\theta}(\mathbf{s}) \cdot \nabla_{\mathbf{a}_j} Q_i^{\mathbf{a}_\theta}(\mathbf{s}, \mathbf{a}_\theta(\mathbf{s}))\right], \quad (7)$$

where to ensure adequate exploration, we apply Ornstein-Uhlenbeck process to add noise on the output of actor network in each time step. Here we further consider the off-policy deterministic actor-critic algorithms (Lillicrap et al. 2015; Silver et al. 2014) to reduce the variance. In particular, we employ a *critic* function in Eq. (7) to estimate the action-value $Q_i^{\mathbf{a}_\theta}$ where off-policy explorations can be conducted. In training the *critic*, we use the sum of square loss and have the following gradient for the parametrised critic $Q^\xi(\mathbf{s}, \mathbf{a})$, where $\xi$ is the parameter for the Q network:

$$\nabla_\xi L(\xi) = \mathbb{E}_{\mathbf{s} \sim \rho^{\mathcal{T}}_{\mathbf{a}_\theta}(\mathbf{s})}\left[\sum_{i=1}^{N} (r_i(\mathbf{s}, \mathbf{a}_\theta(\mathbf{s})) + \lambda Q_i^\xi(\mathbf{s}', \mathbf{a}_\theta(\mathbf{s}'))\right.$$
$$\left. - Q_i^\xi(\mathbf{s}, \mathbf{a}_\theta(\mathbf{s}))) \cdot \nabla_{\partial \xi} Q_i^\xi(\mathbf{s}, \mathbf{a}_\theta(\mathbf{s}))\right]. \quad (8)$$

Note that the gradient is also aggregated from multiple agents as the policy network would do. With Eqs. (7) and Eqs. (8), we apply Stochastic Gradient Descent (SGD) to optimise both the actor and the critic networks. The pseudocode of the over algorithm is given in the *Supplementary Material*.

BiCNet is markedly different from greedy MDP approach as the dependency of agents are embedded in the latent layers, rather than directly on the actions. While simple, our approach allow full dependency among agents because the gradients from all the actions in Eq.(7) are efficiently propagated through the entire networks. Yet, unlike CommNet (Sukhbaatar, Fergus, and others 2016), our communication is not fully symmetric, and we maintain certain social conventions and roles by fixing the order of the agents that join the RNN. This would help solving any possible tie between multiple optimal joint actions (Busoniu, Babuska, and De Schutter 2008; Spaan et al. 2002).

Across different agents, the parameters are shared so that the number of parameters is independent of the number of agents (analogous to the shared parameters across the time domain in vanilla RNN). Parameter sharing results in the compactness of the model which could speed up the learning process. Moreover, this could also enable the domain adaption where the network trained on the small team of of agents (typically three) effectively scales up to larger team of agents during the test under different combat scenarios.

## Experiments

### Experimental Setup

To understand the properties of our proposed BiCNet and its performance, we conducted the experiments on the StarCraft combats with different settings . Following similar experiment set-up as Sukhbaatar, Fergus, and others, BiCNet controls a group of agents trying to defeat the enemy units controlled by the built-in AI.

The level of combat difficulties can be adjusted by varying the unit types and the number of units in both sides. We measured the winning rates, and compared it with the state-of-the-art approaches. The comparative baselines consist of both the rule-based approaches, and deep reinforcement learning approaches. Our setting is summarised as follows where BiCNet controls the former units and the built-in AI controls the latter. We categorize the settings into three types: 1) easy combats {*3 Marines vs. 1 Super Zergling*, and *3 Wraiths vs. 3 Mutalisks*}; 2) hard combats {*5 Marines vs. 5 Marines*, *15 Marines vs. 16 Marines*, *20 Marines vs. 30 Zerglings*, *10 Marines vs. 13 Zerglings*, and *15 Wraiths vs. 17 Wraiths.*}; 3) heterogeneous combats { *2 Dropships and 2 Tanks vs. 1 Ultralisk* }.

The rule-based approaches allow us to have a reference point that we could make sense of. Here we adopted three rule-based baselines: **StarCraft built-in AI**, **Attack the Weakest**, **Attack the Closest**.

For the deep reinforcement learning approaches, we considered the following work as the baselines:

**Independent controller (IND):** We trained the model for single agent and control each agent individually in the combats. Note that there is no information sharing among different agents even though such method is easily adaptable to all kinds of multiagent combats.

**Fully-connected (FC):** We trained the model for all agents in a multiagent setting and control them collectively in the combats. The communication between agents are fully-connected. Note that it is needed to re-train a different model when the number of units at either side changes.

**CommNet:** CommNet (Sukhbaatar, Fergus, and others 2016) is a multiagent network designed to learning to communicate among multiple agents. To make a fair comparison, we implemented both the CommNet and the BiCNet on the same (state, action) spaces and follow the same training processes.

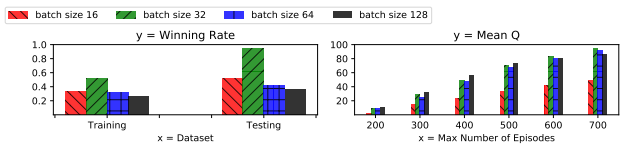

Figure 2: The impact of **batch_size** in combat *2 Marines vs. 1 Super Zergling*.

**GMEZO:** GreedyMDP with Episodic Zero-Order Optimisation (GMEZO) (Usunier et al. 2016) was proposed particularly to solve StarCraft micromanagement tasks. Two novel ideas are introduced: conducting collaborations through a greedy update over MDP agents, as well as adding episodic noises in the parameter space for explorations. To focus on the comparison with these two ideas, we replaced our bi-directional formulation with the greedy MDP approach, and employed episodic zero-order optimisation with noise over the parameter space in the last layer of Q networks in our BiCNet. We keep the rest of the settings exactly the same.

**BiCNet:** In BiCNet, we defined the action space differently from Sukhbaatar, Fergus, and others. Specifically, the action space of each individual agent is represented as a 3-dimensional real vector, *i.e.*, continuous action space. The first dimension corresponds to the probability of attack, according to which we sample a value from [0,1]. If the sampled value is 1, then the agent attacks; otherwise, the agent moves. The second and the third dimension correspond to the degree and the distance of where to attack. With the above three quantities, BiCNet can precisely order an agent to attack a certain location. Note that this is different from executing high-level commands such as 'attack enemy_id', in other words, how to effectively output the damage is itself a form of intelligence.

### Parameter Tuning

In our training, *Adam* (Kingma and Ba 2014) is set as the optimiser with learning rate equal to $0.002$ and the other arguments set by default values. We set the maximum steps of each episode as 800.

We study the impact of the batch size and the results are shown in Figure 2 in the "2 Marines vs. 1 Super Zergling" combat. The two metrics, the winning rate and the Q value, are given. We fine-tune the *batch_size* by selecting the best BiCNet model which are trained on 800 episodes (more than 700k steps) and then tested on 100 independent games. The model with *batch_size* 32 achieves both the highest winning rate and the highest mean Q-value after 600k training steps. We also observed that skip frame 2 gave the highest mean Q-value between 300k and 600k training steps. We fix this parameter with the learned optimal value in the remaining part of our test.

In Fig. 3, we also compare the convergence speed of parameter learning by plotting the winning rate against the number of training episodes. It shows that the proposed BiCNet model has a much quicker convergence than the two main StarCraft baselines.

### Performance Comparison

Table 1 compares our proposed BiCNet model against the baselines under multiple combat scenarios. In each scenario,

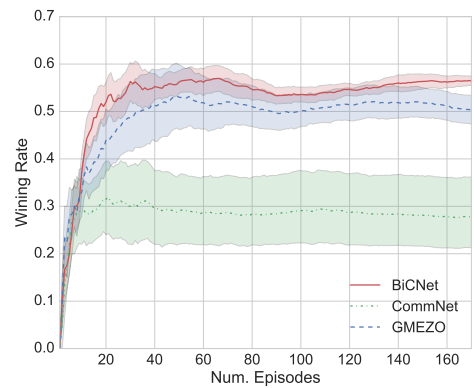

Figure 3: Learning Curves in Combat "10 *Marines* vs. 13 *Zerglings*"

Table 1: Performance comparison. M: *Marine*, Z: *Zergling*, W: *Wraith*.

| Combat | Rule Based | | | RL Based | | | | |
|---|---|---|---|---|---|---|---|---|
| | Built-in | Weakest | Closest | IND | FC | GMEZO | CommNet | BiCNet |
| 20 M vs. 30 Z | **1.00** | .000 | .870 | .940 | .001 | .880 | **1.00** | **1.00** |
| 5 M vs. 5 M | .720 | .900 | .700 | .310 | .080 | .910 | **.950** | .920 |
| 15 M vs. 16 M | .610 | .000 | .670 | .590 | .440 | .630 | .680 | **.710** |
| 10 M vs. 13 Z | .550 | .230 | .410 | .522 | .430 | .570 | .440 | **.640** |
| 15 W vs. 17 W | .440 | .000 | .300 | .310 | .460 | .420 | .470 | **.530** |

BiCNet is trained over 100k steps, and we measure the performance as the average winning rate on 100 test games. The winning rate of the built-in AI is also provided as an indicator of the level of difficulty of the combats.

As illustrated in Table 1, in $4/5$ of the scenarios, BiCNet outperforms the other baseline models. In particular, when the number of agents goes beyond $10$, where cohesive collaborations are required, the margin of the performance gap between BiCNet and the second best starts to increase.

In the combat "5 M vs. 5 M", where the key factor to win is to "focus fire" on the weak, the IND and the FC models have relatively poorer performance. We believe it is because both of the models do not come with an explicit collaboration mechanism between agents in the training stage; coordinating the attacks towards one single enemy is even challenging. On the contrary, GMEZO, CommNet, and BiCNet, which are explicitly or implicitly designed for multiagent collaboration, can grasp and master this simple strategy, thus enjoying better performances. Furthermore, it is worth mentioning that despite the second best performance on "5 Marines vs. 5 Marines", our BiCNet only needs $10$ combats before learning the idea of "focus fire", and achieves over $85\%$ win rate, whereas CommNet needs more than $50$ episodes to grasp the skill of "focus fire" with a much lower winning rate.

Note that the order of which side starts the first attack will influence the combat. This explains why in the combat "5 M vs. 5 M", the built-in AI on the left (as the first to attack) has more advantages on the winning rate $0.720$ over the built-in AI on the right, even though the number of marines at both sides is the same.

### How It Works

To further understand how BiCNet works, we conduct two more studies. We first examine whether a higher Q-value would represent a more optimal join actions among agents.

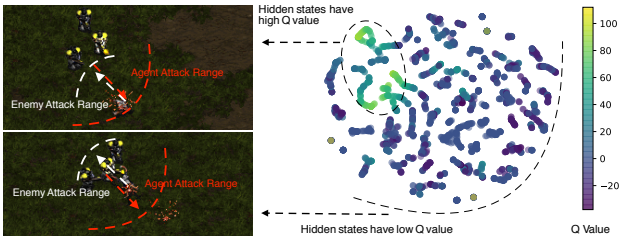

Figure 4: Visualisation for 3 Marines vs. 1 Super Zergling combat. **Upper Left**: State with high Q value; **Lower Left**: State with low Q value; **Right**: Visualisation of hidden layer outputs for each step using TSNE, coloured by Q values.

We visualise the model outputs when the coordinated cover attack is learned in Figure 4. The values in the last hidden layer of the critic network over **10k** steps are collected and then embeded in 2-dimensional space using t-SNE algorithm (Maaten and Hinton 2008). We observe that the steps with high $Q$-values are aggregated in the same area in the embedding space. For example, Figure 4 **Upper Left** shows that the agents attack the enemy in far distance when the enemy cannot attack the agents, and in this status, the model predicts high Q values. By contrast, in Figure 4 **Lower Left**, the agents suffer the damages from the enemy when it closes, which leads to low Q-values.

Our next aim is to examine whether there is any semantic meaning of the information exchanged among agents before their actions. However, due to the high variability of the StarCraft game, so far we have not observed any concrete meaning yet. We instead only focus on bidirectioinal communications by considering a simpler game, where the sophistications that are not related to communications are removed. Specifically, this simpler game consists of $n$ agents, At each round, each agent observes a randomly generated number (sampled in range $[-10, 10]$ under truncated Gaussian) as its input, and nothing else. The goal for each agent is to output the sum over the inputs that all the agents observed. Each agent receives reward based on the difference between the sum and their prediction (action output).

In the setting of three agents guessing the sum with one Bi-RNN communication layer (the hidden state size is 1) followed by a MLP layer, Figure 5 displays the values that have been transferred among three agents. As shown, Agent 1 passes a high value to Agent 2 when it observes a high observation number. When Agent 2 communicates with Agent 3, it tends to output an "additive" value between its own and previously communicated agent, i.e., agent 1. In other words, the hidden state value is increasing when the sum of Agents 1 and 2's observations goes high. Both senders have learned to make the other receiver obtain a helpful message in order to predict the target sum over all agents' observations.

We further set the game with num. of agents $n = 5$, 10, or 20. Apart from the four baselines tested previously, we also implement a supervised MLP with 10 hidden nodes as additional (predicting the sum based on the inputs given to agents). The results are compared in Table 2. The metric is the absolute value of the difference between each agent's action and target. We see our method significantly outperform others. The second best is CommNet. Possible explanation is that it takes an average as the message, and thus naturally fits the problem, while ours have to learn the additives implicitly.

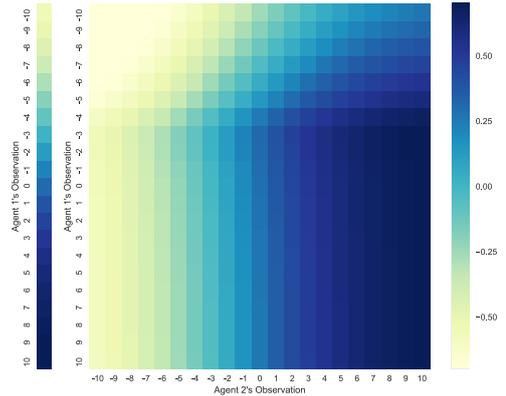

Figure 5: **Left**: The hidden state value passed by Agent 1 to Agent 2 in three agent guessing number game; **Middle**: The hidden state value passed by Agent 1 and Agent 2 to Agent 3 in three agent guessing number game; **Right**: Colour bar.

Table 2: Performance comparison in the guessing game with different agent numbers. Results are given as average $|action\_value - target\_value|$ in $10,000$ testing steps and its standard deviation; A smaller value means a better performance. SL-MLP is a supervised MLP as an additional baseline. t-test is conducted, and the significant ones (p-value $< 0.05$) compared to the second best are marked as *.

| Agent Number | SL-MLP | IND | CommNet | GMEZO | BiCNet |
|---|---|---|---|---|---|
| 5 | 2.82±2.38 | 13.92±12.0 | 0.57±0.41 | 5.92±7.623 | ***0.52**±**0.51** |
| 10 | 4.31±3.67 | 15.32±13.90 | 1.18±0.90 | 9.21±8.22 | ***0.97**±**0.91** |
| 20 | 6.71±5.31 | 19.67±14.61 | 3.88±3.03 | 13.65±11.74 | ***3.12**±**2.93** |

### Emerged Human-level Coordination

With adequate trainings from scratch, BiCNet would be able to discover several effective collaboration strategies. In this section, we conduct a qualitative analysis on the learned collaboration policies from BiCNet. We refer the demonstration video to the *Supplementary Material* and the experimental configurations to Section Experiments.

**Coordinated moves without collision.** We observe that, in the initial stages of learning, in Fig. 6 (a) and (b), the agents move in a rather uncoordinated way. In particular, when two agents are close to each other, one agent often unintentionally blocks the other's path. With the increasing rounds of training (typically over 40k steps in near 50 episodes in the "3

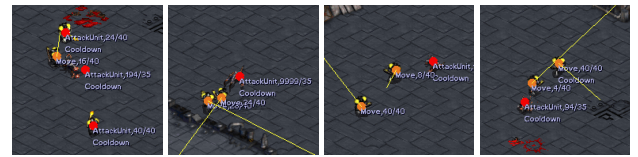

(a) Early stage of training (b) Early stage of training (c) Well-trained (d) Well-trained

Figure 6: Coordinated moves without collision in combat *3 Marines (ours) vs. 1 Super Zergling (enemy)*. The yellow line points out the direction each agent is going to move.

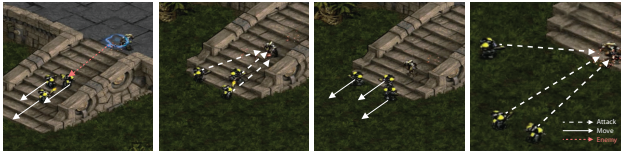

(a) time step 1    (b) time step 2    (c) time step 3    (d) time step 4

Figure 7: *Hit and Run* tactics in combat *3 Marines (ours) vs. 1 Zealot (enemy)*.

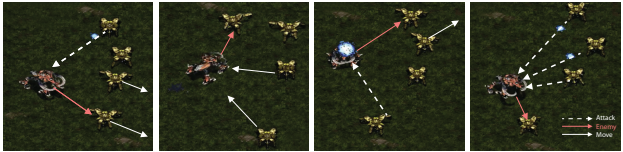

(a) time step 1    (b) time step 2    (c) time step 3    (d) time step 4

Figure 8: Coordinated cover attacks in combat *4 Dragoons (ours) vs. 1 Ultralisks (enemy)*

Marines vs. 1 Super Zergling" combat setting), the number of collisions reduces dramatically. Finally, when the training becomes stable, the coordinated moves emerge, as illustrated in Fig. 6 (c) and (d). Such coordinated moves become important in large-scale combats as shown in Fig. 9 (a) and (b).

**Hit and Run tactics.** For human players, a common tactic of controlling agents in StarCraft combat is *Hit and Run*, i.e., moving the agents away if they are under attack, and fighting back again when agents stay safe. We find that BiCNet can rapidly grasp the tactic of *Hit and Run*, either in the case of single agent or multiple agents settings. We illustrate four consecutive movements of *Hit and Run* in Fig. 7. Despite the simplicity, *Hit and Run* serves as the basis for more advanced and sophisticated collaboration tactics.

**Coordinated cover attack.** *Cover attack* is a high-level collaborative strategy that is often used on the real battlefield. The essence of cover attack is to let one agent draw fire or attentions from the enemies, meanwhile, other agents take advantage of this time period or distance gap to output more harms. The difficulty of conducting cover attack lies in how to arrange the sequential moves of multiple agents in a coordinated *hit and run* way. As shown in Figs. 8, BiCNet can master it well. Starting from Fig. 8(a), BiCNet controls the bottom two *Dragoons* to run away from the enemy *Ultralisk*, while the one in the upper-right corner immediately starts to attack the enemy *Ultralisk* to cover them up. As a response, the enemy starts to attack the top one in time step 2. The bottom two *Dragoons* fight back and form another cover-up. By continuously looping this strategy over, the team of *Dragoons* guarantees consecutive attack outputs to the enemy while minimising the team-level damages (because the enemy wastes time in targeting different *Dragoons*) until the enemy is killed.

**Focus fire without overkill.** As the number of agents increases, how to efficiently allocate the attacking resources becomes important. Neither scattering over all enemies nor focusing on one enemy (wasting attacking fires is also called overkill) are desired. We observe that BiCNet learns to control each agent to focus their fires on particular enemies, and different agents tend to move to the sides to spread the fire and avoid overkill. An example could be found in Fig.(9)

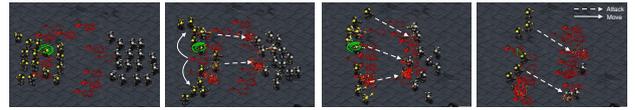

(a) time step 1    (b) time step 2    (c) time step 3    (d) time step 4

Figure 9: "focus fire" in combat *15 Marines (ours) vs. 16 Marines (enemy)*.

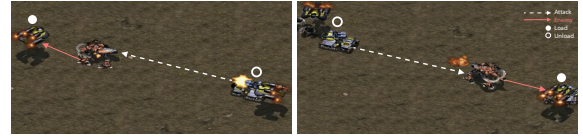

(a) time step 1    (b) time step 2

Figure 10: Coordinated heterogeneous agents in combat *2 Dropships and 2 tanks vs. 1 Ultralisk*.

**Collaborations between heterogeneous agents.** In StarCraft, there are tens of types of agent units, each with unique functionalities, action space, strength, and weakness. For combats with different types of units involved, we expect the agents to reach win-win situations through the collaborations. In fact, heterogeneous collaborations can be easily implemented in our framework by limiting the parameter sharing only to the same types of the units. In this paper, we study a simple case where two *Dropships* and two *tanks* collaborate to fight against an *Ultralisk*. A *Dropship* does not have the function to attack, but it can carry maximally two ground units in the air. As shown in Fig. 10, when the *Ultralisk* is attacking one of the *tanks*, the *Dropship* escorts the *tank* to escape from the attack. At the same time, the other *Dropship* unloads his *tank* to the ground so as to attack the *Ultralisk*. At each side, the collaboration between the *Dropship* and the *tank* keeps iterating until the *Ultralisk* is destroyed.

## Conclusions

In this paper, we have introduced a new deep multiagent reinforcement learning. The action is learned by constructing a vectorised actor-critic framework, where each dimension corresponds to an agent. The coordination is done by bidirectional recurrent communications in the internal layers. Through end-to-end learning, our BiCNet would be able to successfully learn several effective coordination strategies. Our experiments have demonstrated its ability to collaborate and master diverse combats in StarCraft combat games. We have also shown five human-level coordination strategies BiCNet could grasp from playing StarCraft combat games. Admittedly, quantifying the sophistication of the collaborations in games is challenging in general, and our analysis here is qualitative in nature.

In the next step, we plan to carry on experiments of letting the machine compete with human players at different levels. We also plan to further investigate how the policies are communicated over the networks among agents in more complicated settings, and whether there is a specific language that may have emerged in StartCraft (Lazaridou, Peysakhovich, and Baroni 2016; Mordatch and Abbeel 2017).

# Supplementary Material

**Proof of Theorem 1**

Following the regularity conditions mentioned in (Silver et al. 2014), we know that the supreme of $\frac{\partial Q_i^{\mathbf{a}_\theta}(\mathbf{s},\mathbf{a})|_{\mathbf{a}=\mathbf{a}_\theta(\mathbf{s})}}{\partial \mathbf{a}_i}$ and $\frac{\partial \mathbf{a}_{i,\theta}(\mathbf{s})}{\partial \theta}$ for each agent $i$ are bounded functions of $\mathbf{s}$. Based on the regularity and the boundedness, we can use Leibniz integral rule and Fubini's theorem, respectively. Note that as the policy $\mathbf{a}_\theta$ and the transition matrix of the environment $\mathcal{T}$ are both considered deterministic, the expectation is only taken over the initial state $\mathbf{s}_0$, which is different from the original deterministic policy gradient theorem. According to the definition of $Q_i^{\mathbf{a}_\theta}(\mathbf{s},\mathbf{a})$ and the our objective function in Eq.(6), we derive the multiagent deterministic policy gradient theorem, which mostly follows the line of (Sutton et al. 2000).

$$\frac{\partial J(\theta)}{\partial \theta} = \frac{\partial}{\partial \theta} \int_{\mathcal{S}} p_1(\mathbf{s}) \sum_{i=1}^{N} Q_i^{\mathbf{a}_\theta}(\mathbf{s}, \mathbf{a}_\theta(\mathbf{s})) d\mathbf{s} \tag{9}$$

$$= \int_{\mathcal{S}} p_1(\mathbf{s}) \frac{\partial}{\partial \theta} \sum_{i=1}^{N} Q_i^{\mathbf{a}_\theta}(\mathbf{s}, \mathbf{a}_\theta(\mathbf{s})) d\mathbf{s} \tag{10}$$

$$= \int_{\mathcal{S}} p_1(\mathbf{s}) \frac{\partial}{\partial \theta} \sum_{i=1}^{N} \left( r_i(\mathbf{s}, \mathbf{a}_\theta(\mathbf{s})) + \int_{\mathcal{S}} \lambda \mathbb{1}(\mathbf{s}' = \mathcal{T}^1_{\mathbf{a}_\theta, \mathbf{b}_\phi}(\mathbf{s})) Q_i^{\mathbf{a}_\theta}(\mathbf{s}', \mathbf{a}_\theta(\mathbf{s}')) d\mathbf{s}' \right) d\mathbf{s} \tag{11}$$

$$= \int_{\mathcal{S}} p_1(\mathbf{s}) \left( \frac{\partial \mathbf{a}_\theta(\mathbf{s})}{\partial \theta} \frac{\partial}{\partial \mathbf{a}} \sum_{i=1}^{N} r_i(\mathbf{s}, \mathbf{a})|_{\mathbf{a}=\mathbf{a}_\theta(\mathbf{s})} \right) d\mathbf{s} +$$
$$\int_{\mathcal{S}} p_1(\mathbf{s}) \int_{\mathcal{S}} \lambda \left( \frac{\partial \mathbf{a}_\theta(\mathbf{s})}{\partial \theta} \frac{\partial}{\partial \mathbf{a}} \mathbb{1}(\mathbf{s}' = \mathcal{T}^1_{\mathbf{a}, \mathbf{b}_\phi}(\mathbf{s}))|_{\mathbf{a}=\mathbf{a}_\theta} \cdot \sum_{i=1}^{N} Q_i^{\mathbf{a}_\theta}(\mathbf{s}', \mathbf{a}_\theta(\mathbf{s}')) \right) d\mathbf{s}' d\mathbf{s} +$$
$$\int_{\mathcal{S}} p_1(\mathbf{s}) \int_{\mathcal{S}} \lambda \left( \mathbb{1}(\mathbf{s}' = \mathcal{T}^1_{\mathbf{a}_\theta, \mathbf{b}_\phi}(\mathbf{s})) \cdot \frac{\partial}{\partial \theta} \sum_{i=1}^{N} Q_i^{\mathbf{a}_\theta}(\mathbf{s}', \mathbf{a}_\theta(\mathbf{s}')) \right) d\mathbf{s}' d\mathbf{s} \tag{12}$$

$$= \int_{\mathcal{S}} p_1(\mathbf{s}) \left( \frac{\partial \mathbf{a}_\theta(\mathbf{s})}{\partial \theta} \frac{\partial}{\partial \mathbf{a}} \sum_{i=1}^{N} Q_i^{\mathbf{a}_\theta}(\mathbf{s}, \mathbf{a})|_{\mathbf{a}=\mathbf{a}_\theta(\mathbf{s})} + \int_{\mathcal{S}} \lambda \left( \mathbb{1}(\mathbf{s}' = \mathcal{T}^1_{\mathbf{a}_\theta, \mathbf{b}_\phi}(\mathbf{s})) \cdot \underbrace{\frac{\partial}{\partial \theta} \sum_{i=1}^{N} Q_i^{\mathbf{a}_\theta}(\mathbf{s}', \mathbf{a}_\theta(\mathbf{s}'))}_{\text{iterate as Eq.(10) to Eq.(11)}} \right) d\mathbf{s}' \right) d\mathbf{s} \tag{13}$$

$$= \int_{\mathcal{S}} \int_{\mathcal{S}} \sum_{t=0}^{\infty} \lambda^t p_1(\mathbf{s}) \mathbb{1}(\mathbf{s}' = \mathcal{T}^t_{\mathbf{a}_\theta, \mathbf{b}_\phi}(\mathbf{s})) \frac{\partial \mathbf{a}_\theta(\mathbf{s}')}{\partial \theta} \frac{\partial}{\partial \mathbf{a}} \sum_{i=1}^{N} Q_i^{\mathbf{a}_\theta}(\mathbf{s}', \mathbf{a})|_{\mathbf{a}=\mathbf{a}_\theta(\mathbf{s}')} d\mathbf{s}' d\mathbf{s} \tag{14}$$

$$= \int_{\mathcal{S}} \underbrace{\left( \int_{\mathcal{S}} \sum_{t=0}^{\infty} \lambda^t p_1(\mathbf{s}) \mathbb{1}(\mathbf{s}' = \mathcal{T}^t_{\mathbf{a}_\theta, \mathbf{b}_\phi}(\mathbf{s})) d\mathbf{s} \right)}_{\rho^{\mathcal{T}}_{\mathbf{a}_\theta}(\mathbf{s}')} \frac{\partial \mathbf{a}_\theta(\mathbf{s}')}{\partial \theta} \frac{\partial}{\partial \mathbf{a}} \sum_{i=1}^{N} Q_i^{\mathbf{a}_\theta}(\mathbf{s}', \mathbf{a})|_{\mathbf{a}=\mathbf{a}_\theta(\mathbf{s}')} d\mathbf{s}' \tag{15}$$

$$= \mathbb{E}_{\mathbf{s} \sim \rho^{\mathcal{T}}_{\mathbf{a}_\theta}(\mathbf{s})} \left[ \frac{\partial \mathbf{a}_\theta(\mathbf{s}')}{\partial \theta} \frac{\partial}{\partial \mathbf{a}} \sum_{i=1}^{N} Q_i^{\mathbf{a}_\theta}(\mathbf{s}', \mathbf{a})|_{\mathbf{a}=\mathbf{a}_\theta(\mathbf{s}')} \right] \tag{16}$$

$$= \mathbb{E}_{\mathbf{s} \sim \rho^{\mathcal{T}}_{\mathbf{a}_\theta}(\mathbf{s})} \left[ \sum_{i=1}^{N} \sum_{j=1}^{N} \nabla_\theta \mathbf{a}_{j,\theta}(\mathbf{s}) \cdot \nabla_{\mathbf{a_j}} Q_i^{\mathbf{a}_\theta}(\mathbf{s}, \mathbf{a}_\theta(\mathbf{s})) \right], \tag{17}$$

where in Eq.(10) Leibniz intergal rule is used to exchange derivative and integral since $Q_i^{\mathbf{a}_\theta}(\mathbf{s}, \mathbf{a}_\theta(\mathbf{s}))$ is continuous. For Eq.(11), we used the definition of $Q$-value. Then, we take derivatives for each term in Eq.(11) to get Eq.(12). Afterwards, we combine the first and the second term in Eq.(12) to get the first term in Eq.(13), while we notice that we can iterate Eq.(10) and Eq.(11) to expand the second term in Eq.(13). By summing up the iterated terms, we get Eq.(14), which implies Eq.(15) by using Fubini's theorem to exchange the order of integration. Using the expectation to denote Eq.(15), we derive Eq.(16). Finally, we get Eq.(17) and the proof is done.

## Pseudocode

**Algorithm 1** BiCNet algorithm

Initialise actor network and critic network with $\xi$ and $\theta$
Initialise target network and critic network with $\xi' \leftarrow \xi$ and $\theta' \leftarrow \theta$
Initialise replay buffer R
**for** episodes=1, E **do**
  initialise a random process $\mathcal{U}$ for action exploration
  receive initial observation state $s^1$
  **for** t=1, T **do**
    for each agent $i$, select and execute action $a_i^t = \mathbf{a}_{i,\theta}(s^t) + \mathcal{N}_t$
    receive reward $[r_i^t]_{i=1}^N$ and observe new state $s^{t+1}$
    store transition $\{s^t, [a_i^t, r_i^t]_{i=1}^N, s^{t+1}\}$ in R
    sample a random minibatch of $M$ transitions $\{s_m^t, [a_{m,i}^t, r_{m,i}^t]_{i=1}^N, s_m^{t+1}\}_{m=1}^M$ from R
    compute target value for each agent in each transition using the Bi-RNN:
    **for** m=1, M **do**
$$\hat{Q}_{m,i} = r_{m,i} + \lambda Q_{m,i}^{\xi'}(s_m^{t+1}, \mathbf{a}_{\theta'}(s_m^{t+1})) \text{ for each agent } i$$
    **end for**
    compute critic gradient estimation according to Eq.(8):
$$\Delta\xi = \frac{1}{M}\sum_{m=1}^M \sum_{i=1}^N \left[(\hat{Q}_{m,i} - Q_{m,i}^{\xi}(\mathbf{s}_m, \mathbf{a}_\theta(\mathbf{s}_m))) \cdot \nabla_\xi Q_{m,i}^{\xi}(\mathbf{s}_m, \mathbf{a}_\theta(\mathbf{s}_m))\right].$$
    compute actor gradient estimation according to Eq.(7) and replace $Q$-value with the critic estimation:
$$\Delta\theta = \frac{1}{M}\sum_{m=1}^M \sum_{i=1}^N \sum_{j=1}^N \left[\nabla_\theta \mathbf{a}_{j,\theta}(\mathbf{s}_m) \cdot \nabla_{\mathbf{a_j}} Q_{m,i}^{\xi}(\mathbf{s}_m, \mathbf{a}_\theta(\mathbf{s}_m))\right]$$
    update the networks based on Adam using the above gradient estimators
    update the target networks:
$$\xi' \leftarrow \gamma\xi + (1-\gamma)\xi', \qquad \theta' \leftarrow \gamma\theta + (1-\gamma)\theta'$$
  **end for**
**end for**